\documentclass[runningheads,a4paper]{llncs}

\usepackage{amssymb}
\usepackage{graphicx,bm,array}

\usepackage{url}
\urldef{\mailsa}\path|{w.j.wilkinson@qmul.ac.uk}|
\newcommand{\keywords}[1]{\par\addvspace\baselineskip
\noindent\keywordname\enspace\ignorespaces#1}

\begin{document}

\mainmatter  

\title{A Generative Model for Natural Sounds Based on Latent Force Modelling}

\titlerunning{Latent Force Modelling of Natural Sounds}

%
%
\author{William J. Wilkinson
\and Joshua D. Reiss\and Dan Stowell}
\authorrunning{Latent Force Modelling of Natural Sounds}

\institute{Centre for Digital Music,\\
Queen Mary University of London\\
\mailsa\\
}

%
%

\toctitle{A Generative Model for Natural Sounds}
\tocauthor{William J. Wilkinson}
\maketitle

\begin{abstract}
Recent advances in analysis of subband amplitude envelopes of natural sounds have resulted in convincing synthesis, showing subband amplitudes to be a crucial component of perception. Probabilistic latent variable analysis is particularly revealing, but existing approaches don't incorporate prior knowledge about the physical behaviour of amplitude envelopes, such as exponential decay and feedback. We use latent force modelling, a probabilistic learning paradigm that incorporates physical knowledge into Gaussian process regression, to model correlation across spectral subband envelopes. We augment the standard latent force model approach by explicitly modelling correlations over multiple time steps. Incorporating this prior knowledge strengthens the interpretation of the latent functions as the source that generated the signal. We examine this interpretation via an experiment which shows that sounds generated by sampling from our probabilistic model are perceived to be more realistic than those generated by similar models based on nonnegative matrix factorisation, even in cases where our model is outperformed from a reconstruction error perspective.
\end{abstract}

\section{Introduction}
Computational models for generating audio signals are a means of exploring and understanding our perception of sound. Natural sounds, defined here as everyday non-music, non-speech sounds, are an appealing medium with which to study perception since they exclude cognitive factors such as language and musical interpretation. McDermott \cite{mcdermott2011sound} used synthesis as a means to demonstrate that the human auditory system utilises time-averaged statistics of subband amplitudes to classify sound textures. In a similar vein, Turner \cite{turner2010statistical} constructed a synthesis model based on probabilistic latent variable analysis of those same subband amplitudes. One main advantage of a latent variable approach is the possibility that the uncovered latent behaviour may represent either \emph{i)} the primitive source that generated the signal, or \emph{ii)} the latent information that the human auditory system encodes when it calculates time-averaged statistics.

Latent variable analysis captures correlations across multiple dimensions by modelling the data's shared dependence on some unobserved (latent) variable or function. It is, by its very nature, ill-posed; we typically aim to simultaneously predict both the latent functions \emph{and} the mapping from this latent space to the observation data. As such, infinitely many potential solutions exist and we cannot guarantee that our prediction will encode the true sound source or our true perceptual representation.

The ill-posed nature of the problem necessitates the use of prior information. It is commonly suggested that nonnegativity, smoothness and sparsity form a suitable set of prior assumptions about real life signals. We argue that, even after imposing such constraints, a simple scalar mapping between the latent space and observation space is insufficient to capture all the complex behaviour that we observe in the subband amplitude envelopes of an audio signal. We construct a latent force model (LFM) \cite{alvarez2009latent} to incorporate prior knowledge about how amplitude envelopes behave via a discrete differential equation that models exponential decay \cite{wilkinson2017latent}.



Utilising the state space formulation \cite{hartikainen2011sequential}, we augment the standard LFM by explicitly including in the current state information from many discrete time steps. This allows us to capture phenomena such as feedback, damping and to some extent reverberation. In this probabilistic approach the latent functions are modelled with Gaussian processes, which provide uncertainty information about our predictions whilst also guaranteeing that the latent functions are smooth. Nonnegativity is imposed via a nonlinear transformation.

Evaluating latent representations is not straightforward. Objective measures of our ability to reconstruct the observation data don't inform us about the interpretability of our predictions. We hypothesise that if the latent functions capture physically or perceptually meaningful information, then a generative model based on synthesising latent functions that are statistically similar should generate realistic data when projected back to the observation space.

In this paper we introduce a generative model, applicable to a wide range of natural sounds, based on an extended LFM\footnote{Matlab source code and example stimuli can be found at \protect\url{c4dm.eecs.qmul.ac.uk/audioengineering/natural_sound_generation}} (Section \ref{sec:LFMaudio}). Comparative models based on variants of nonnegative matrix factorisation (NMF) are implemented to perform evaluation-by-synthesis, which shows how listeners often perceive the LFM approach to generate more realistic sounds even in cases where NMF is more efficient from a reconstruction error perspective (Section \ref{sec:results}).

\vspace{-0.2cm}

\section{Background}
\label{sec:background}

\vspace{-0.1cm}

The perceptual similarity of two sounds is not determined by direct comparison of their waveforms, but rather by comparison of their statistics \cite{mcdermott2011sound}. Hence it is argued that prior information for natural sounds should take a statistical form \cite{turner2010statistical}. We argue in Section \ref{sec:LFMaudio} that these statistical representations can be improved through the inclusion of assumptions about the physical behaviour of sound, resulting in a hybrid statistical-physical prior.

In order to analyse sound statistics, both McDermott \cite{mcdermott2011sound} and Turner \cite{turner2010statistical} utilise the subband filtering approach to time-frequency analysis, in which the signal is split into different frequency channels by a bank of band-pass filters. The time-frequency representation is then formed by tracking the amplitude envelopes of each subband. McDermott generates sound textures by designing an objective function which allows the statistics of a synthetic signal to be matched to that of a target signal. Turner utilises probabilistic time-frequency analysis combined with probabilistic latent variable analysis to represent similar features. Turner's approach has the advantage that once the parameters have been optimised, new amplitude envelopes can be generated by drawing samples from the latent distribution. It should be noted that samples drawn from the model will not exhibit the fast attack and slow decay we observe in audio amplitude envelopes, since the model is temporally symmetric.

NMF is a ubiquitous technique for decomposing time-frequency audio data \cite{fevotte2009nonnegative,smaragdis2003non,virtanen2007monaural}, however a common criticism is its inability to take into account temporal information. The most common approach to dealing with this issue is to impose smoothness on the latent functions, the idea being that smoothness is a proxy for local correlation across neighbouring time steps. Temporal NMF (tNMF) imposes smoothness by penalising latent functions which change abruptly \cite{virtanen2007monaural} or by placing a Gaussian process prior over them \cite{turner2014time}. An alternative approach is to use a hidden Markov model to capture the changes in an audio signal's spectral make up over time \cite{mysore2010non}. High resolution NMF (HR-NMF) models the temporal evolution of a sound by utilising the assumption that natural signals are a sum of exponentially modulated sinusoids, with each frequency channel being assigned its own decay parameter estimated using expectation-maximisation \cite{badeau2011gaussian}. 

\vspace{-0.2cm}

\subsection{Latent Force Models}
\label{ssec:LFMs}

To incorporate our prior assumptions into data-driven analysis we use latent force models (LFMs) \cite{alvarez2009latent}, a probabilistic modelling approach which assumes $M$ observed output functions $x_m$ are produced by some $R<M$ unobserved (latent) functions $u_r$ being passed through a set of differential equations. If the chosen set of differential equations represents some physical behaviour present in the system we are modelling, even if only in a simplistic manner, then such a technique can improve our ability to learn from data \cite{alvarez2013linear}. This is achieved by placing a Gaussian process (GP) prior \cite{rasmussen2006gaussian} over the $R$ latent functions, calculating the cross-covariances (which involves solving the ODEs), and performing regression.



It was shown by Hartikainen and S\"arkka \cite{hartikainen2011sequential} that, under certain conditions, an equivalent regression task can be performed by reformulating the model (i.e. the ODE representing our physical knowledge of the system) into state space (SS) form, reformulating the GP as a stochastic differential equation (SDE), and then combining them into a joint SS model:
\begin{equation} \label{joint}
\frac{{d\bm{{x}}(t)}}{dt}=\bm{f}(\bm{{x}}(t))+Lw(t) \;  .
\end{equation}
Here $\bm{{x}}(t)$ represents the state vector containing $\{x_m(t)\}^M_{m=1}$ \emph{and} the states of the SDE $\{u_r(t),\dot{u}_r(t),...\}^R_{r=1}$, $w(t)$ is a white noise process, $\bm{f}$ is the transition function which is dependent on $\theta$, the set of all ODE parameters and GP / SDE hyperparameters, and $L$ is a vector determining which states are driven by the white noise. The model's discrete form is
\begin{equation} \label{discModel}
\bm{{x}}[t_k]=\bm{\hat{f}}(\bm{{x}}[t_{k-1}],\Delta t_{k})+\bm{q}[t_{k-1}] \;  ,
\end{equation}
where $\Delta_t$ is the time step size, $\bm{\hat{f}}$ is the discretised transition function and $\bm{q}[t_{k-1}]\sim N(\bm{{0}},Q[\Delta t_{k}])$ is the noise term with process noise matrix $Q$. The corresponding output measurement model is
\begin{equation} \label{measurement}
\bm{y}[t_k] = H\bm{x}[t_k] + \epsilon[t_k], \hspace{0.3cm} \epsilon[t_k] \sim N(0,\sigma^2) \;  ,
\end{equation}
where measurement matrix $H$ simply selects the outputs from the joint model.

The posterior process $\bm{x}[t_k]$, i.e. the solution to (\ref{discModel}), is a GP in the linear case such that the filtering distribution $p(\bm{x}[t_{k}]\hspace{0.1cm}|\hspace{0.1cm}\bm{y}[t_1],...,\bm{y}[t_{k}])$ is Gaussian. Hence state estimation can be performed via Kalman filtering and smoothing \cite{sarkka2013bayesian}. 

However, if $\bm{f}$ is a nonlinear function, as is the case if we wish to impose nonnegativity on the latent functions, then calculation of the predictive and filtering distributions involves integrating equations which are a combination of Gaussian processes and nonlinear functions. We may approximate the solutions to these integrals numerically using Gaussian cubature rules. This approach is known as the cubature Kalman filter (CKF) \cite{hartikainen2012state}.

The Kalman update steps provide us with the means to calculate the marginal data likelihood $p(\bm{y}[t_{1:T}]\hspace{0.1cm}|\hspace{0.1cm}\theta)$. Model parameters $\theta$ can therefore be estimated from the data by maximising this likelihood using gradient-based methods.

\vspace{-0.2cm}

\section{Latent Force Models for Audio Signals}
\label{sec:LFMaudio}

\vspace{-0.2cm}


To obtain amplitude data in the desired form we pass an audio signal through an equivalent rectangular bandwidth (ERB) filter bank. We then use Gaussian process probabilistic amplitude demodulation (GPPAD) \cite{turner2011demodulation} to calculate the subband envelopes and their corresponding carrier signals. GPPAD allows for control over demodulation time-scales via GP lengthscale hyperparameters. We are concerned with slowly varying behaviour correlated across the frequency spectrum, in accordance with the observation that the human auditory system summarises sound statistics over time \cite{mcdermott2011sound}. Fast-varying behaviour is relegated to the carrier signal and will be modelled as independent filtered noise.

The number of channels in the filter bank and the demodulation lengthscales must be set manually during this first analysis stage. Keeping the number of total model parameters small is a priority (see Section \ref{ssec:AmpEnvLFM}), so we typically set the number of filters to 16, and the lengthscales such that we capture amplitude behaviour occurring over durations of 10ms and slower. 



\vspace{-0.2cm}

\subsection{Augmented Latent Force Models for Amplitude Envelopes}
\label{ssec:AmpEnvLFM}

We use a first order differential equation to model the exponential decay that occurs in audio amplitude envelopes \cite{wilkinson2017latent}. However this overly simplistic model does not take into account varying decay behaviour due to internal damping, or feedback and other nonstationary effects which occur as a sound is generated and propagates towards a listener.

Since we require nonnegativity of our latent functions, which is imposed via nonlinear transformation, we use the nonlinear LFM whose general from is (\ref{discModel}) with nonlinear $\bm{\hat{f}}$. For a first order ODE its discrete form is
\begin{equation} \label{eq:LFM}
\dot{x}_m[t_k]=-D_mx_m[t_k]+\sum^R_{r=1} S_{mr}g(u_r[t_k]) \;  ,
\end{equation}
for $m=1,...,M$ where $M$ is the number of frequency channels. $D_m$ and $S_{mr}$ are the damping and sensitivity parameters respectively and $g(u)=log(1+e^u)$ is the positivity-enforcing nonlinear transformation. The model progresses forwards in time with step size $\Delta_t$ using Euler's method: ${x}_m[t_{k+1}]={x}_m[t_k] + \Delta_t\dot{x}_m[t_k]$.

To account for the complex behaviour mentioned above that occurs in real audio signals, we extend this discrete model such that predictions at the current time step $t_k$ can be influenced explicitly by predictions from multiple time steps in the past. As in \cite{wilkinson2017latent} we augment the model by adding a parameter $\gamma_m$ which controls the ``linearity" of decay. Our final model becomes
\begin{equation} \label{myModel}
\dot{x}_m[t_k]=-D_mx^{\gamma_m}_m[t_k]+\sum_{p=1}^PB_{mp}x_m[t_{k-p}]+\sum_{q=0}^P\sum^R_{r=1} S_{mrq}g(u_r[t_{k-q}]) \; .
\end{equation}
We restrict $\gamma_m \in [0.5,1]$, and for sounding objects with strong internal damping we expect $\gamma_m$ to be small, representing an almost linear decay. Parameters $B_{mp}$ are \emph{feedback} coefficients which determine how the current output is affected by output behaviour from $p$ time steps in the past. $S_{mrq}$ are \emph{lag} parameters which determine how sensitive the current output is to input $r$ from $q$ time steps ago.

The lag term is important since modes of vibration in a sounding object tend to be activated at slightly different times due to deformations in the object as it vibrates, and due to the interaction of multiple modes of vibration. It can also capture effects due to reverberation. The feedback terms allow for long and varied decay behaviour that can't be described by simple exponential decay.

\begin{figure}[!t]
\centering
\includegraphics[width=8cm]{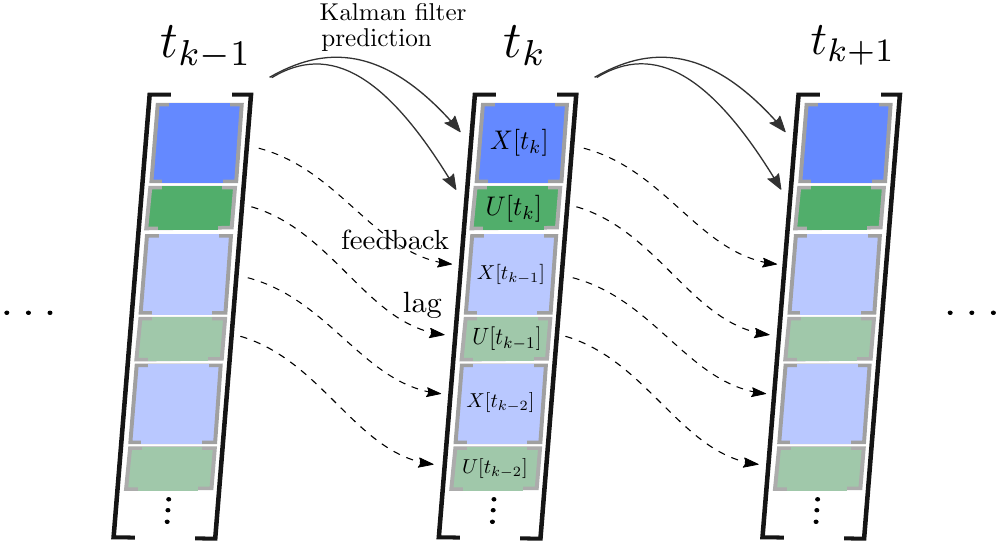}
\vspace{-0.3cm}
\caption{The augmented LFM stores terms from previous time steps in the state vector. Blue represents output predictions $X$ (amplitudes), green represents latent predictions $U$. Each step, predictions pass down to feedback and lag state locations. The entire state is used to predict the next step's outputs and latents via Kalman filtering.}
\label{fig:lfm_diagram}
\vspace{-0.3cm}
\end{figure}

The challenge is to incorporate (\ref{myModel}) into our filtering procedure. We do this by augmenting our state vector $\bm{{x}}[t_k]$ and transition model
\begin{equation} \label{eq:transitionModel}
\bm{\hat{f}}(\bm{{x}}[t_{k-1}],\Delta t_{k}) = \bm{{x}}[t_k] + \Delta_t\bm{\dot{x}}[t_k]
\end{equation}
with new rows corresponding to the delayed terms. Fig. \ref{fig:lfm_diagram} shows how after each time step the current states $X[t_k]=\{x_m[t_k]\}_{m=1}^M, U[t_k]=\{u_r[t_k]\}_{r=1}^R$  are ``passed down'' such that at the next time step they are in the locations corresponding to feedback and lag terms. When performing the Kalman filter prediction step, augmented states are included since they influence predictions for the current state, however the predictions for these augmented entries are simply exact copies from the previous time step.

\begin{figure}[t]
\centering
\includegraphics[width=12cm]{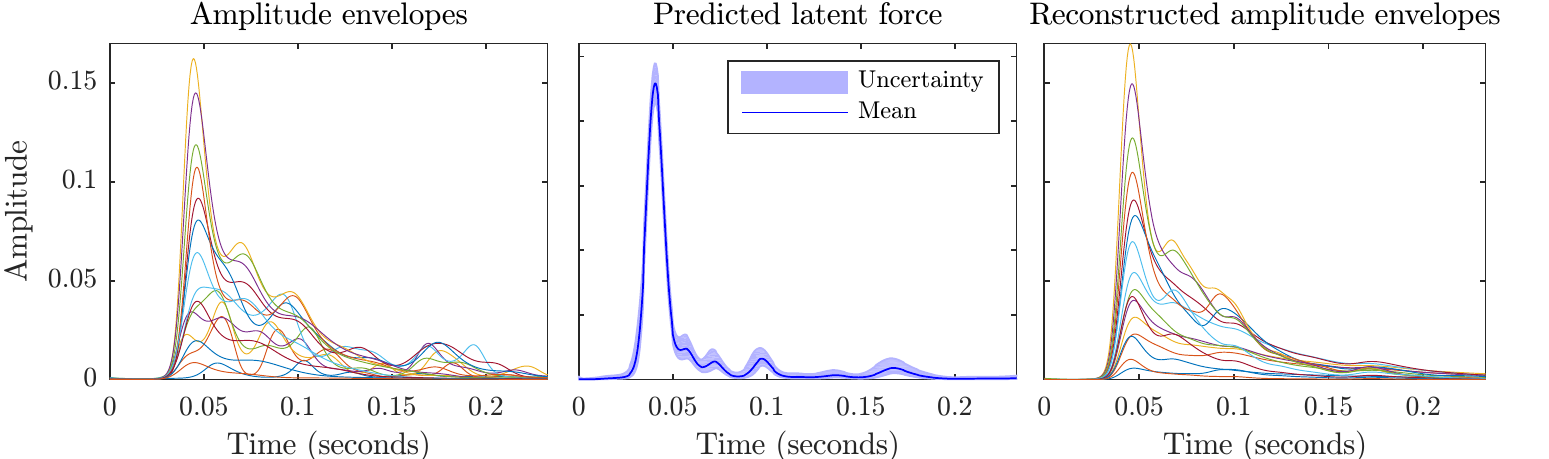}
\vspace{-0.3cm}
\caption{LFM applied to a metal impact sound, with mean and 95\% confidence of the latent distribution shown. The mean is passed through the model (\ref{myModel}) to reconstruct the envelopes. Complex behaviour is maintained despite using a single force.}
\label{fig:lfmMetal}
\vspace{-0.3cm}
\end{figure}

Fig. \ref{fig:lfmMetal} shows the latent prediction for a metal impact sound with one latent force, $R=1$. The mean of the distribution is the minimum least squares error estimate, so we pass it through discrete model (\ref{myModel}) to reconstruct the amplitude envelopes. Despite the single latent force, we observe that some of the complex behaviour has been learnt. Additionally, the latent force is both smooth and sparse, and the reconstructed envelopes have a slow decay despite this sparsity.

\vspace{-0.2cm}

\subsection{Generating Novel Instances of Natural Sounds}
\label{ssec:genModel}

A significant benefit of probabilistic approaches such as LFM or tNMF is that, as well as providing us with uncertainty information about our predictions, they provide the means to sample new latent functions from the learnt distribution. By passing these new functions through the model we can generate amplitude envelopes. These envelopes modulate carrier signals produced using a sinusoids-plus-noise approach based on analysis of the original carriers. The subbands are then summed to create a new synthetic audio signal distinct from the original but with similar characteristics.

\begin{figure}[t]
\centering
\includegraphics[width=11.25cm]{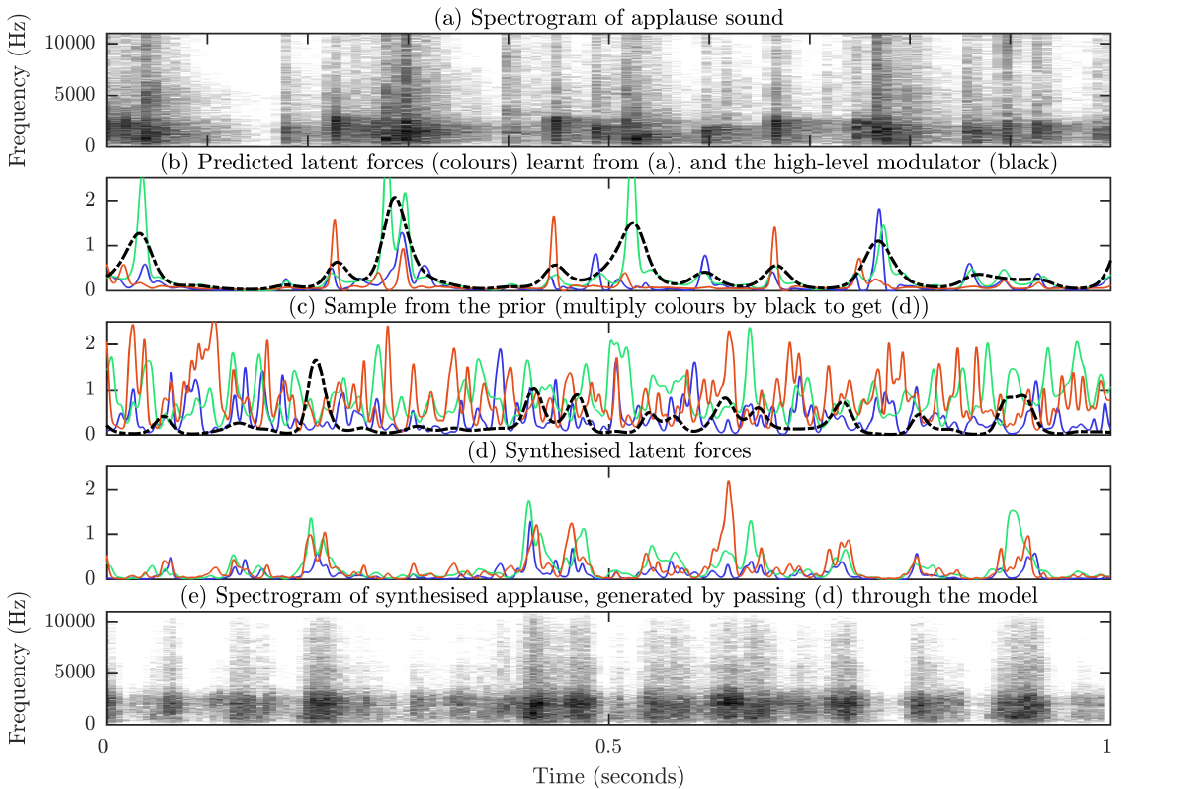}
\vspace{-0.3cm}
\caption{LFM generative model with 3 latent forces applied to an applause sound. The high-level modulator (black line in (b)) is calculated by demodulating the latent forces.}
\label{fig:genModel}
\vspace{-0.4cm}
\end{figure}

Sampling from the prior of the learnt distribution generates functions with appropriate smoothness and magnitude, however the desired energy sparsity is not guaranteed. Latent functions are modelled independently, but in practice they tend to co-occur and are activated in similar regions of the signal. We use GPPAD again to demodulate our latent functions with a slowly varying envelope, then fit a GP with a squared exponential covariance function to this envelope \cite{rasmussen2006gaussian}. We sample from this high-level envelope and use it to modulate our newly generated latent functions; the results of this product is latent behaviour with sparse energy, as demonstrated in Fig. \ref{fig:genModel}(d).

\vspace{-0.2cm}

\subsection{Optimisation Settings}
\label{ssec:paramOpt}

The set of model parameters $\{D_{m},B_{mp},S_{mrq},\gamma_m,\lambda_r\}$, with GP lengthscales $\lambda_r$, becomes large as $R$, $P$ increase. To alleviate issues that occur when our parameter space becomes large we sparsify the feedback and sensitivity parameters. For example, if $P=10$, we may manually fix $B_{mp}$ to zero for $p\in[3,4,6,7,9]$ such that only half the parameters are included in the optimisation procedure.

Reliability of the optimisation procedure suffers as the number of parameters increases, so in practice all $M$ frequency channels are not optimised together. We select the $6$ envelopes contributing the most energy and train the model on the observations from only these channels. The remaining channels are then appended on and optimised whilst keeping the already-trained parameters fixed. This improves reliability but prioritises envelopes of high energy. We also skip prediction steps for periods of the signal that are of very low amplitude, which speeds up the filtering step. Despite these adjustments, optimisation still takes up to 3 days for a 2 second sound sample.



\section{Evaluation}
\label{sec:results}

To evaluate our method we collated a set of 20 audio recordings, selected as being representative of everyday natural sounds\footnote{From \protect{\url{freesound.org}} and from the Natural Sound Stimulus set: \protect\url{mcdermottlab.mit.edu/svnh/Natural-Sound/Stimuli.html}}. Music and speech sounds were not included, nor were sounds with significant frequency modulation, since our model doesn't capture this behaviour.

\vspace{-0.2cm}

\subsection{Reconstruction Error of Original Sound}
\label{ssec:reconError}

\begin{figure}[!t]
\centering
\includegraphics[width=12cm]{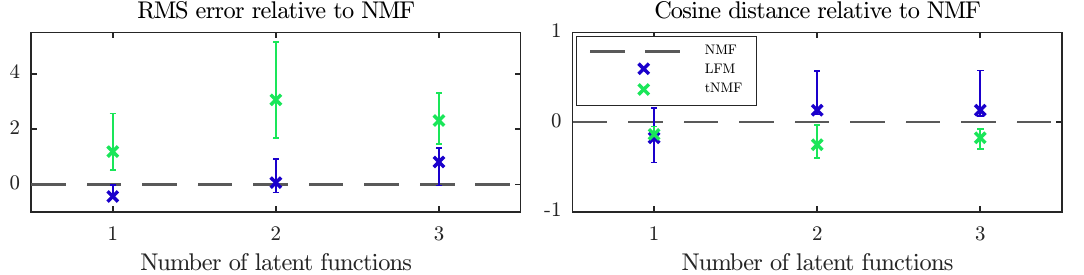}
\vspace{-0.3cm}
\caption{Reconstruction error of LFM and tNMF plotted relative to NMF. Crosses represent the median, error bars range from first to third quartile.}
\label{fig:objRes}
\vspace{-0.3cm}
\end{figure}

We analyse our ability to reconstruct the original data by projecting the latent representation back to the output space. For the LFM this means passing the mean of the learnt distribution through model (\ref{myModel}).  Fig. \ref{fig:objRes} shows reconstruction RMS error and cosine distance of LFM and tNMF relative to NMF for the 20 recordings. The smoothness constraint enforced by placing a GP prior over the latent functions negatively impacts the reconstruction. This is demonstrated by the fact that tNMF performs poorly from an RMS error perspective. Despite this, the LFM has much descriptive power, and is sometimes capable of achieving a lower RMS error than the unconstrained NMF. Interestingly however, tNMF consistently outperforms the other two models based on cosine distance.

\subsection{Listening Test for Novel Sounds}
\label{ssec:listTest}

Objective results suggest that smoothness constraints harm reconstruction of the original signal. However, our aim is to learn realistic latent representations that will be the foundation of a generative model. To test their suitability, we designed an experiment to compare generative models based on LFM, NMF and tNMF. The approach outlined in Section \ref{ssec:genModel} was used for all model types. Since NMF is non-probabilistic, it does not provide an immediate way in which to sample new data, therefore GPs were fit to the latent functions after analysis.

\begin{figure}[t]
\centering
\includegraphics[width=12cm]{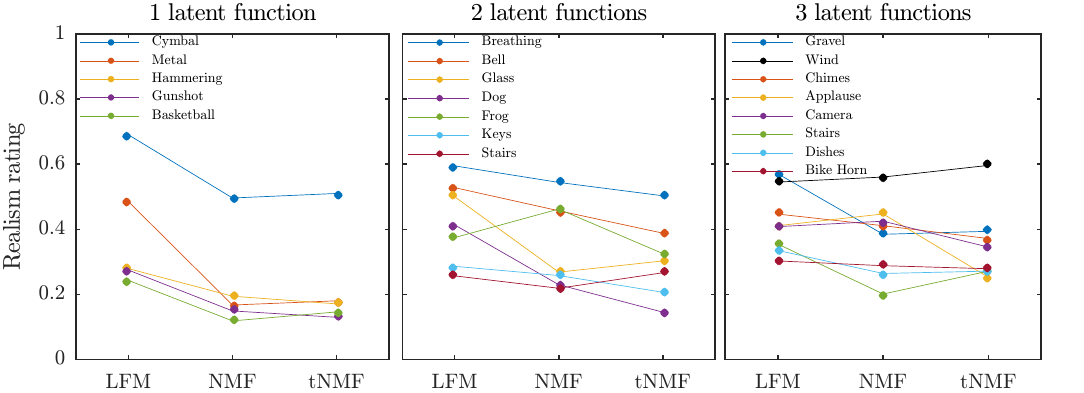}
\vspace{-0.3cm}
\caption{Mean realism ratings obtained from the listening test.}
\label{fig:listTest}
\vspace{-0.3cm}
\end{figure}

Our experiment followed a multi-stimulus subjective quality rating paradigm\footnote{The test was run online and implemented with the Web Audio Evaluation Tool: \protect{\url{github.com/BrechtDeMan/WebAudioEvaluationTool}}}: 24 participants were shown 20 pages (order randomised), one per sound example, and asked to listen to the reference recording and then rate 7 generated sounds (2 from each model plus an anchor) based on their credibility as a new sound of the same type as the reference. Ratings were on a scale of 0 to 1, with a score of 1 representing a very realistic sound. Fig. \ref{fig:listTest} shows the mean realism ratings. Whilst variation was large between sound examples, LFM was generally rated as more realistic than the other methods.

To test for significance we applied a generalised linear mixed effects model (GLMM), with beta regression, in which \emph{sound example} and \emph{participant} were treated as random effects. Table 1 shows that the mean realism rating was highest for LFM regardless of number of latent functions. The difference was significant at a 5\% level except for LFM vs. NMF with 3 latent functions. This suggests that for sounds requiring many latent functions to capture their behaviour, such as textural sounds, LFM may not offer a significant gain over purely statistical approaches. For example, the wind recording in Fig. \ref{fig:listTest}, a textural sound whose envelopes do not exhibit clear exponential decay, was captured best with tNMF.


\vspace{-0.4cm}

\newcolumntype{P}[1]{>{\centering\arraybackslash}p{#1}}
\begin{table}[!ht]
\fontsize{8}{8.2}\selectfont
\setlength\extrarowheight{3pt}
\begin{center}
\begin{tabular} { |p{2.0cm}||P{1.15cm} P{1.15cm}|P{1.15cm} P{1.15cm}|P{1.15cm} P{1.15cm}|P{1.15cm} P{1.15cm}| } 
 \hline
  & \multicolumn{2}{c}{All sounds} & \multicolumn{2}{c}{1 latent fn.} & \multicolumn{2}{c}{2 latent fns.} & \multicolumn{2}{c|}{3 latent fns.} \\
 \hline
  & Estimate & p value & Estimate & p value & Estimate & p value & Estimate & p value\\
 \hline
 LFM vs. NMF & 0.3839 & \textbf{\textless1e-04} & 0.8248 & \textbf{\textless1e-05}& 0.3140 & \textbf{0.0448} & 0.2052 & 0.2867\\
 
 LFM vs. tNMF & 0.4987 & \textbf{\textless1e-04} & 0.7976 & \textbf{\textless1e-05} & 0.5134 & \textbf{\textless0.001} & 0.3243 & \textbf{0.0285}\\
 
 NMF vs. tNMF & 0.1148 & 0.3750 & -0.0272 & 0.9980 & 0.1994 & 0.3218 & 0.1191 & 0.7154\\
 \hline
\end{tabular}
\vspace{0.1cm}
\caption{\label{AllTable}{GLMM with three-way comparison applied to listening test results. LFM received higher mean ratings, but confidence decreases with number of latent forces, indicated by increasing \emph{p values}. \emph{Estimate} can be interpreted as the ratio increase in realism rating when choosing model A over model B.}}
\end{center}
\end{table}

\vspace{-1.3cm}

\section{Conclusion}
\label{sec:conclusion}

\vspace{-0.2cm}

Our results show that in order to extend existing synthesis techniques to a larger class of sounds, it is important to utilise prior knowledge about how natural sound behaves. We achieved this by using latent force modelling to capture exponential decay, and augmented the standard approach to include feedback and delay across many discrete time steps. Doing so allowed us to make smooth, sparse latent predictions that we argue are more representative of the real source that generated a given sound.

This claim is supported by the fact that a generative model based on LFM was consistently rated as more realistic by listeners than alternatives based on variants of NMF, even in cases where it was not superior in reconstruction of the original signal. Resonance, decay and modulations in the subband amplitudes were captured well by our model, which is flexible enough to be applicable to sounds ranging from glass breaking to dogs barking.

The nonlinear ODE representing our physical knowledge contains a large number of parameters, making our approach impractical in some cases, so a more compact model would be of huge benefit. Efficient nonlinear filtering methods or numerical ODE solvers would make the computation time more acceptable. Future work includes amplitude behaviour occurring on multiple time scales at once, and models for frequency modulation and other nonstationary effects would further expand the class of sounds to which such techniques can be applied.

\vspace{-0.3cm}

\bibliographystyle{splncs}
\bibliography{LVA18refs.bib}

\end{document}